\documentclass[runningheads]{llncs}
\usepackage[T1]{fontenc}
\usepackage{graphicx}
\usepackage{amsmath}
\usepackage{hyperref}
\usepackage{xcolor}  % Required for color
\usepackage{CJKutf8}
\usepackage{paralist}

\begin{document}
\title{Factual Inconsistencies in Multilingual Wikipedia Tables}
\author{Silvia Cappa\inst{1}\and
Lingxiao Kong\inst{2}\and
Pille-Riin Peet\inst{3}\and
Fanfu Wei\inst{4} \and
Yuchen Zhou\inst{5} \and
Jan-Christoph Kalo\inst{6}
}

\institute{ 
CNR ISTC
\email{silviacappa@cnr.it}\\
\and
Fraunhofer Institute for Applied Information Technology FIT
\email{lingxiao.kong@fit.fraunhofer.de}\\
\and
Tallinn University of Technology
\email{pille-riin.peet@taltech.ee}\\
\and
EURECOM
\email{fanfu.wei@eurecom.fr}\\
\and
Technical University of Munich
\email{yuchen.zhou@tum.de}\\
\and
University of Amsterdam
\email{j.c.kalo@uva.nl}\\
}
\maketitle              % typeset the header of the contribution
\begin{abstract}
%The abstract should briefly summarize the contents of the paper in 150--250 words.
Wikipedia serves as a globally accessible knowledge source with content in over 300 languages. Despite covering the same topics, the different versions of Wikipedia are written and updated independently. This leads to factual inconsistencies that can impact the neutrality and reliability of the encyclopedia and AI systems, which often rely on Wikipedia as a main training source. This study investigates cross-lingual inconsistencies in Wikipedia's structured content, with a focus on tabular data. We developed a methodology to collect, align, and analyze tables from Wikipedia multilingual articles, defining categories of inconsistency. We apply various quantitative and qualitative metrics to assess multilingual alignment using a sample dataset.
%Our results show that...
These insights have implications for factual verification, multilingual knowledge interaction, and design for reliable AI systems leveraging Wikipedia content.

\end{abstract}

\section{Introduction }
% background and challenges
Wikipedia is one of the most widely used public knowledge sources, offering content in over 300 languages \cite{enwiki:1294528760}. While articles in different language editions often aim to describe the same entities and events, they frequently diverge in the facts and contents they present, since they are not simply translated but rather compiled independently based on each language version~\cite{tatariya2024good}. These inconsistencies raise important questions about the reliability, completeness, richness, and neutrality of multilingual content. 
This project investigates various inconsistencies across Wikipedia language editions, with a specific focus on structured data such as tables. 
Inconsistency between Wikipedia tables is a heavily understudied problem.
While there are first works on matching and finding incomplete Wikipedia infoboxes in various language versions~\cite{khincha-etal-2025-leveraging} recently, inconsistencies in tables have not been explored at all.

\begin{figure}[t]
    \centering
    \includegraphics[width=0.95\linewidth]{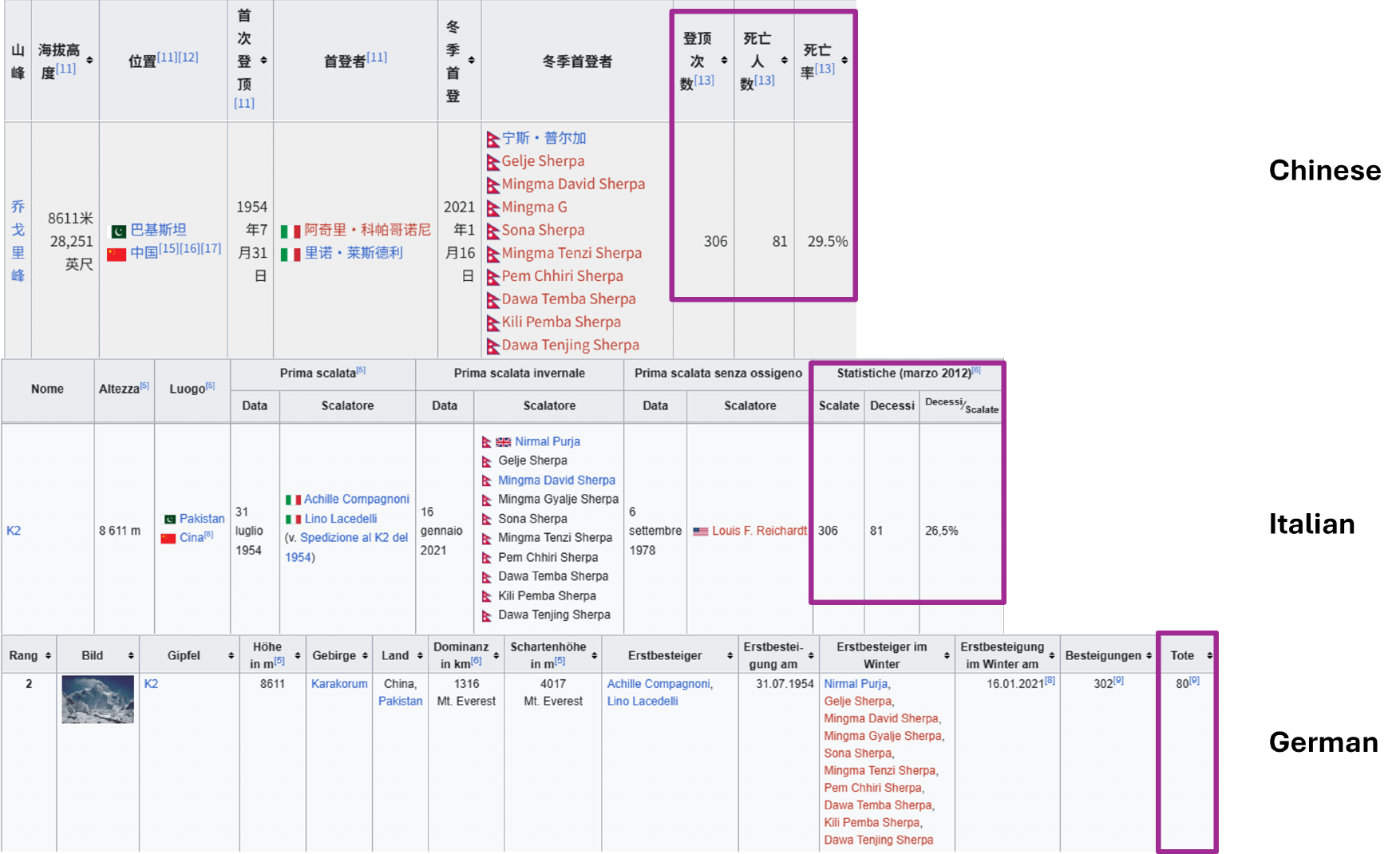}
    \caption{Inconsistencies in death rate information across language versions of the Wikipedia articles about the Seven Summits in Chinese, Italian, and German.}
    \label{fig:teaser-figure}
\end{figure}

As an example of these inconsistencies, in Figure~\ref{fig:teaser-figure}, we show an example of tables about the Seven Summits in 3 different languages. 
While these tables contain the same seven mountains, they have a different structure and are inconsistent.
We aim to categorize these inconsistencies, understand their causes, and assess their implications for knowledge quality and multilingual AI applications.
%There is no precise count of the number of tables in Wikipedia, as they are inserted dynamically within articles and are not a fundamental element of the site's architecture. However, Wikipedia makes extensive use of tables to present organized data, and their presence is widespread across various articles and pages \cite{} what??.

% NPOV policy how it can apply for lists/ranking of entities?

\section{Research Questions}
\label{sec:rq}
In this work, we explore three central research questions:
\begin{enumerate}
    \item \textit{What kinds of factual inconsistencies occur across Wikipedia language editions?}  
    Multilingual tables about the same entity may diverge in the facts they present, the structure of attributes, or the presence of information. These inconsistencies include contradictions, omissions, and misaligned values.

    \item \textit{Can these inconsistencies be categorized into meaningful types?}  
    We aim to define a taxonomy of inconsistency types—such as omissions, contradictions, structural differences, and partial overlaps—to support systematic analysis across languages.

    \item \textit{What are the underlying causes of these inconsistencies?}  
    We investigate whether editorial practices, asynchronous updates, cultural framing, or language-specific conventions contribute to the observed divergences.
\end{enumerate}

% Understanding these causes is crucial for improving the reliability and coherence of multilingual knowledge sources.

%\begin{enumerate}[label=\textbf{RQ\arabic*.},leftmargin=1.1cm]
%\item What kinds of factual inconsistencies exist between different Wikipedia language editions?
%\item Can we define useful categories of inconsistency, for example, omissions, contradictions, unaligned standards or partial overlap?
%\item What might explain these inconsistencies, such as editorial differences, missing updates, or linguistic variation? %look in discussion pages
%\end{enumerate}

\section{Knowledge Graphs for Reliable AI}

KGs provide structured representations of entities and their relations, offering a foundation for consistent and explainable AI. In multilingual contexts, they serve as a common reference across languages, reducing ambiguity and supporting alignment. Wikidata, for example, enables cross-lingual linking of Wikipedia content, including tables, through shared identifiers. This makes it possible to detect and analyze inconsistencies in Wikipedia tables. Wikipedia remains one of the most widely used sources for pretraining large language models, making data quality a central concern in reliable AI \cite{albalak2024survey}. By grounding extracted facts in a knowledge graph, we can improve the reliability of downstream tasks such as question answering, entity linking, or summarization, and increase the robustness of AI systems that rely on multilingual and collaborative sources like Wikipedia.

\section{Related Works }
% \paragraph{Bias and Inconsistency in NLP}
Bias and inconsistency in knowledge sources are well-documented challenges. A comprehensive survey shows how definitions of bias influence methods and evaluations in NLP \cite{blodgett2020language}. Cultural and linguistic bias has been observed in Wikipedia and Wikidata. For example, comparative studies of biographies highlight framing differences across language editions \cite{callahan2011cultural}, while demographic attributes such as nationality or ethnicity are modeled inconsistently in Wikidata \cite{shaik2021analyzing}.

Existing work has addressed these issues through automated bias detection in Wikipedia articles \cite{hube2018detecting} and curated probes to test cultural knowledge in language models \cite{keleg2023dlama}. However, structured content such as tables remains largely unstudied, despite their central role in many Wikipedia articles and the high variability across language editions. Efforts to unify Wikipedia content, such as Abstract Wikipedia \cite{vrandecic2020architecture,vrandecic2021building}, introduce a language-independent layer, but do not account for inconsistencies in existing articles or enable cross-lingual comparison of structured data. To date, no large-scale effort has examined how Wikipedia tables diverge across language editions.
% This project addresses this gap by analyzing cross-lingual inconsistencies in Wikipedia tables.

As a related effort for assessing knowledge consistency, \cite{xing2024evaluating} implemented Cross-lingual Semantic Consistency (xSC) metrics and examined whether models provide semantically consistent responses to the same Wikipedia information in different languages, using multilingual semantic encoding models like LASER. Recent approaches leverage deep learning models and natural language processing techniques to parse table schemas, extract entity relationships, and convert tabular content into knowledge graphs or structured databases for downstream applications~\cite{10.1007/978-3-030-30793-6_21,10.1007/978-3-030-62419-4_20,10.1145/3442442.3458611}. 

While prior work has focused on synchronizing Wikipedia Infoboxes using LLMs to enrich or update missing information \cite{khincha-etal-2025-leveraging}, our study investigates factual inconsistencies in already existing multilingual Wikipedia tables. 
In contrast to Wikipedia Infoboxes, tables have an even larger structural variety, because they often describe many entities, while Infoboxes usually concern the main entity of the respective Wikipedia article.
This adds a lot of additional new challenges.

%\begin{align}
%\text{xSC} &= \frac{1}{L(L-1)} \sum_{i=1}^{L} \sum_{\substack{j=1 \\ %j \neq i}}^{L} \mathbf{C}_{i,j} \tag{1} \\
%\mathbf{C}_{i,j} &= \frac{1}{N} \sum_{s=1}^{N} \text{Cos}%(\text{emb}_s^i, \text{emb}_s^j) \\
%\text{emb}_s^i &= \text{LaBSE}(\text{ans}_s^i)
%\end{align}

% \textbf{Table Fact Extraction}

% %Related work on Table Fact Extraction
% \cite{10.1007/978-3-030-30793-6_21}
% \cite{10.1007/978-3-030-62419-4_20}
% \cite{10.1145/3442442.3458611}

\section{Methodology}

\begin{figure}[t]
    \centering
    \includegraphics[width=0.85\linewidth]{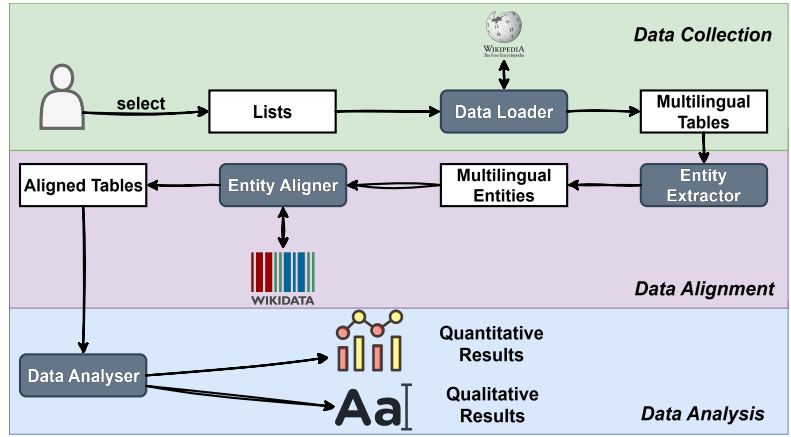}
    \caption{Methodology Overview}
    \label{fig:methodology}
\end{figure}

Figure~\ref{fig:methodology} presents an overview of our proposed approach. To address our research questions, we adopt a data-driven methodology with three steps:

\begin{enumerate}
    \item \textbf{Data Collection} \\
    Our overarching goal is to analyze the content of multilingual Wikipedia pages to study cross-lingual inconsistencies in factual information. As a first step toward this objective, we focus on tabular data, which often provides high-density, structured information such as statistics, factual lists, and attributes of related entities. Tables also present a manageable starting point for aligning content across languages due to their structural format. To this end, we manually select 10 entities from the Wikipedia \textit{List of lists of lists}\footnote{\url{https://en.wikipedia.org/wiki/List_of_lists_of_lists}}, and for each entity, we retrieve the full set of tables from its Wikipedia page in multiple languages using a custom Python script. 
    
    % The selection is based on the following criteria:
    % \begin{itemize}
    %     \item[\textit{i.}] To ensure consistency and illustrative clarity, all entities are chosen from a single domain—\textit{Geography and Places}.
    %     \item[\textit{ii.}] For multilingual applicability, each entity must have Wikipedia pages in more than three languages.
    %     \item[\textit{iii.}] Since our focus is on tabular data, each entity must contain at least one table on its Wikipedia page.
    % \end{itemize}

    \item \textbf{Data Alignment} \\
    After collecting the tables, we perform entity-level processing to align tabular content across languages, which includes entity extraction and entity linking:
    \begin{itemize}
        \item \textbf{Entity Extraction} \\
        Each extracted table typically contains information about multiple entities (e.g., mountains, rivers, lakes). To prepare for cross-lingual comparison, we identify the entity represented in each row, usually based on the name or link in the first column. These entity mentions are then extracted for subsequent alignment across languages.

        \item \textbf{Entity Linking} \\
        \begin{CJK*}{UTF8}{gbsn}
        To unify entity mentions across languages, we leverage Wikidata IDs as a language-independent identifier. For each row-level entity mention, we extract the internal Wikipedia link and query the MediaWiki API to resolve its corresponding Wikidata QID. This allows us to associate mentions like \textit{Mount Everest} (English), \textit{珠穆朗玛峰} (Simplified Chinese), \textit{Il monte Everest} (Italian) with the same unique identifier \textit{Q513}.
        \end{CJK*}
    \end{itemize}

    \item \textbf{Data Analysis} \\
    Following alignment, we perform a quantitative and qualitative assessment of inconsistencies in tabular data across languages. The dataset and specific evaluation metrics used in our analysis are detailed in Section~\ref{sec:datasetandmetrics}.
\end{enumerate}

% domain (data) selection, criteria for inconsistencies categorization, what metrics for evaluation

\section{Dataset and Metrics}
\label{sec:datasetandmetrics}
The goal of the data collection process is to obtain high-quality multilingual tabular data from Wikipedia pages. To ensure a comprehensive evaluation of multilingual inconsistencies, we manually select entities exclusively from the \textit{Geography} domain for consistency and clarity.

% we establish key criteria for selecting appropriate Wikipedia pages and formulate three guiding principles for identifying suitable pages:

% \begin{itemize}
%     \item \textbf{Language sufficiency:} Each entity must have Wikipedia pages in over ten languages.

%     \item \textbf{Domain constraint:} Entities are restricted to the \textit{Geography and Places} domain for consistency and clarity.

%     \item \textbf{Table inclusion:} Each entity’s Wikipedia page must contain at least one table.
% \end{itemize}

As shown in Table~\ref{tab:language_versions}, basic statistics of the dataset demonstrate the number of language editions available for selected Wikipedia articles related to geographical and geological topics, revealing substantial variation in multilingual coverage ranging from 6 to 58 languages. For comparison, we extract English, German, Chinese, Italian, and Dutch versions, which represent widely used languages.

\begin{table}[h!]
\centering
\caption{Number of language versions for selected Wikipedia articles}
\label{tab:language_versions}
\begin{tabular}{|p{8.5cm}|c|}
\hline
\textbf{Title} & \textbf{Language Versions} \\
\hline
Seven Summits & 58 \\
Eight-thousander & 57 \\
List of mountains of the Alps over 4000 metres & 15 \\
Lists of earthquakes & 38 \\
List of highest unclimbed peaks & 6 \\
List of highest mountains on Earth & 47 \\
Lakes of Titan & 18 \\
List of largest lakes of Europe & 18 \\
List of lakes by area & 46 \\
\hline
\end{tabular}
\end{table}

% Inconsistency metrics:
% \begin{itemize}
%     \item incompleteness: missing datapoints
%     \item differences in table sizes
%     \item unaligned standards (metrics)
%     \item number of references
% \end{itemize}

% We can classify the inconsistency categories as followings:

\begin{inparaenum}[(1)]
For evaluation, we utilize various quantitative metrics to assess the collected dataset: 
\item \textbf{Table count}: Although Wikipedia pages describe the same entities, they employ different numbers of tables to elaborate on them. 
\item \textbf{Reference count}: Editors attach references at the end of pages, indicating their level of engagement in providing content. 
\item \textbf{Column count}: This provides a straightforward way to observe how much detail editors include about entities, as some columns may contain missing cells. 
\end{inparaenum}

\begin{inparaenum}[(1)]
Additionally, we qualitatively analyze the following metrics: 
\item \textbf{Invalidity}: The provided value is incorrect or not credible. 
\item \textbf{Timeliness}: A data source may present information that was once valid but is now outdated, whereas another source may provide an updated version.
\item \textbf{Incompleteness}: Schema-level incompleteness, where different language versions provide different information to describe the underlying subject.
\end{inparaenum}

\section{Experiments}

To address the research questions defined in Section~\ref{sec:rq}, we conduct a series of pilot experiments following the methodology outlined earlier. We prepare a small-sized dataset and perform data alignment. After storing the aligned data, we analyze it using both qualitative and quantitative methods based on the defined evaluation metrics. 
The quantitative analysis provides comprehensive statistics about the dataset, as detailed in Section~\ref{sec:quantitative-analysis}. In the qualitative analysis, we categorize the observed inconsistencies and provide specific examples for each inconsistency type, which are described in Section~\ref{sec:qualitative-analysis}.

\subsection{Quantitative Analysis}
\label{sec:quantitative-analysis}
The quantitative analysis examines Wikipedia's multilingual content consistency using three key metrics: table count, reference count, and column count. The findings reveal significant disparities in content coverage and quality:

\begin{figure}[t!]
    \centering
    \includegraphics[width=0.80\linewidth]{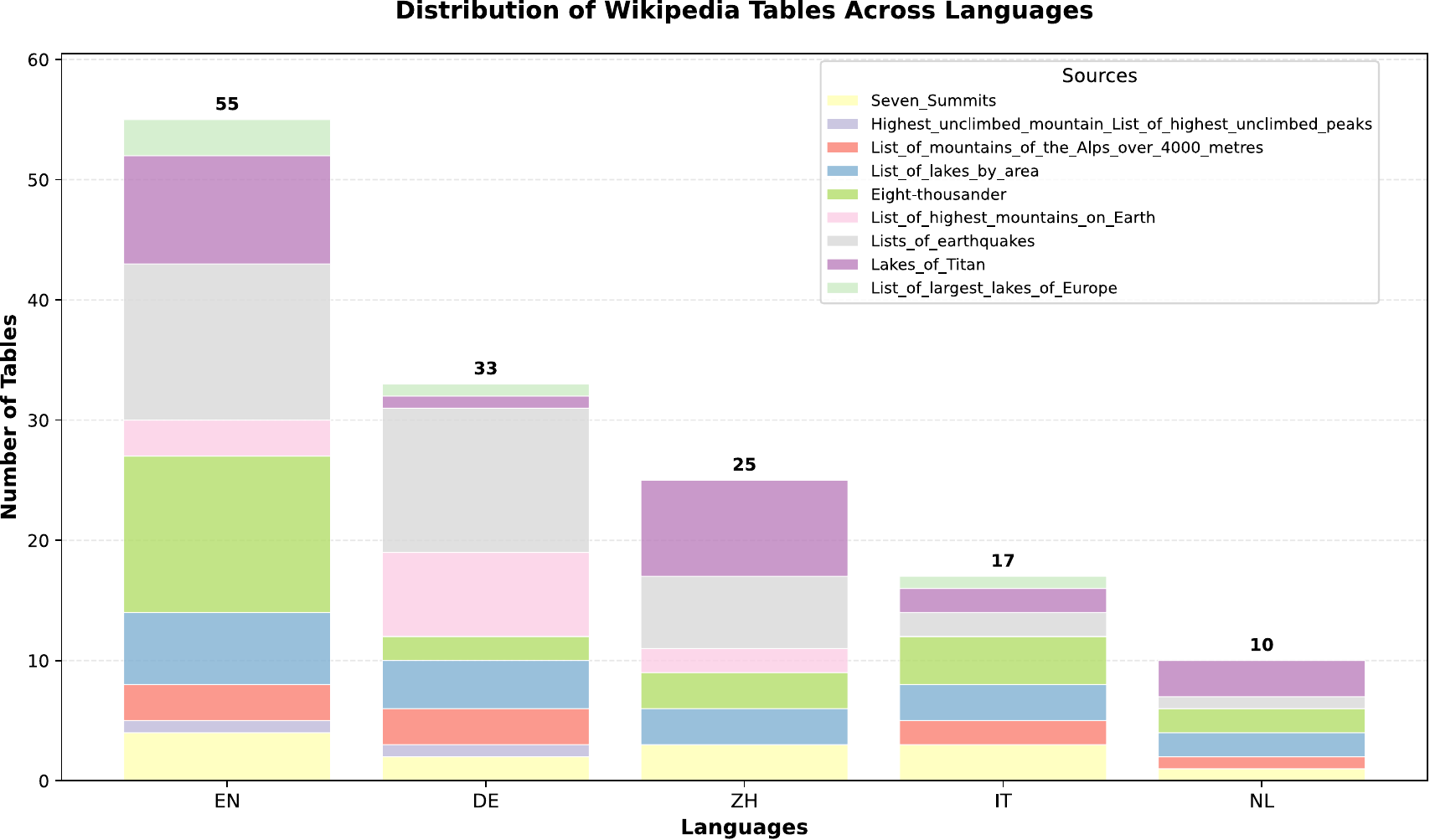}
    \caption{Distribution of Table Numbers across Languages}
    \label{fig:table_numbers}
\end{figure}

\begin{figure}[t!]
    \centering
    \includegraphics[width=0.80\linewidth]{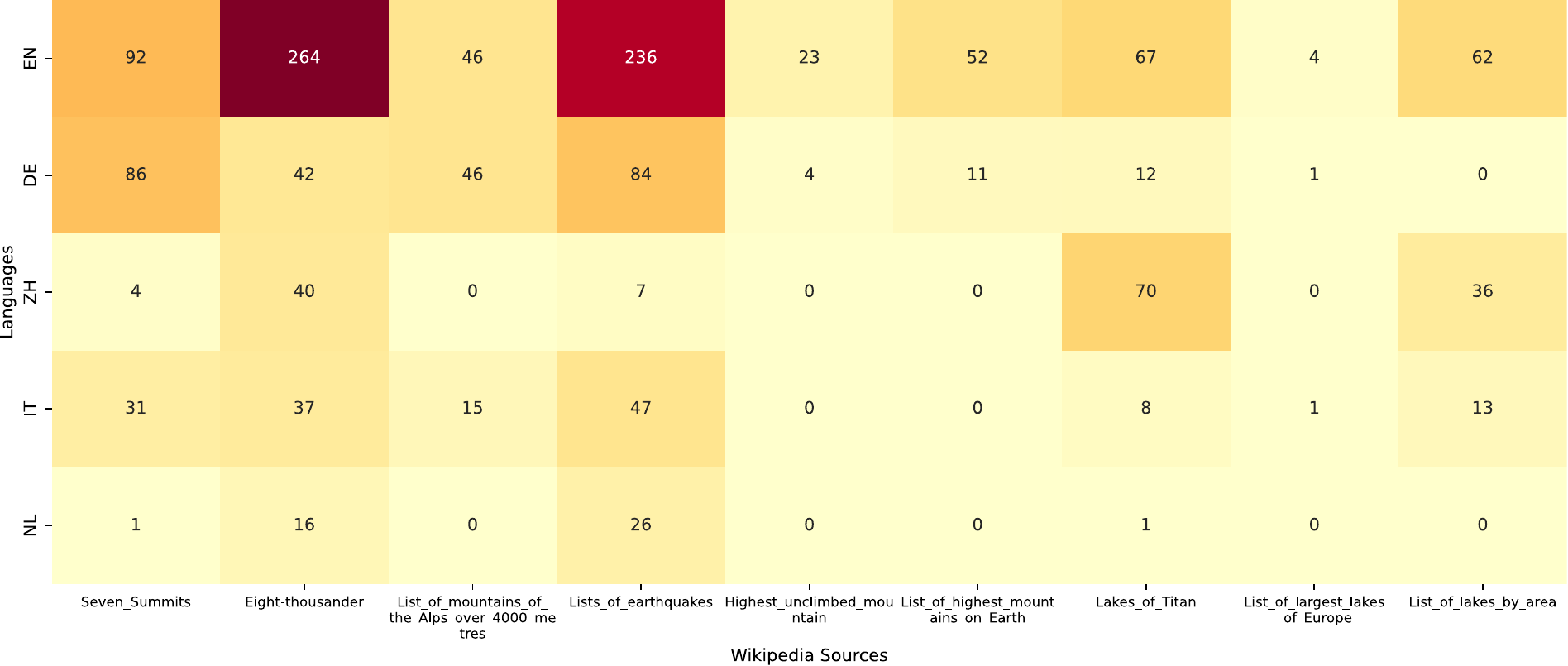}
    \caption{Distribution of Reference Numbers across Languages}
    \label{fig:reference_numbers}
\end{figure}

\begin{figure}[t!]
    \centering
    \includegraphics[width=0.80\linewidth]{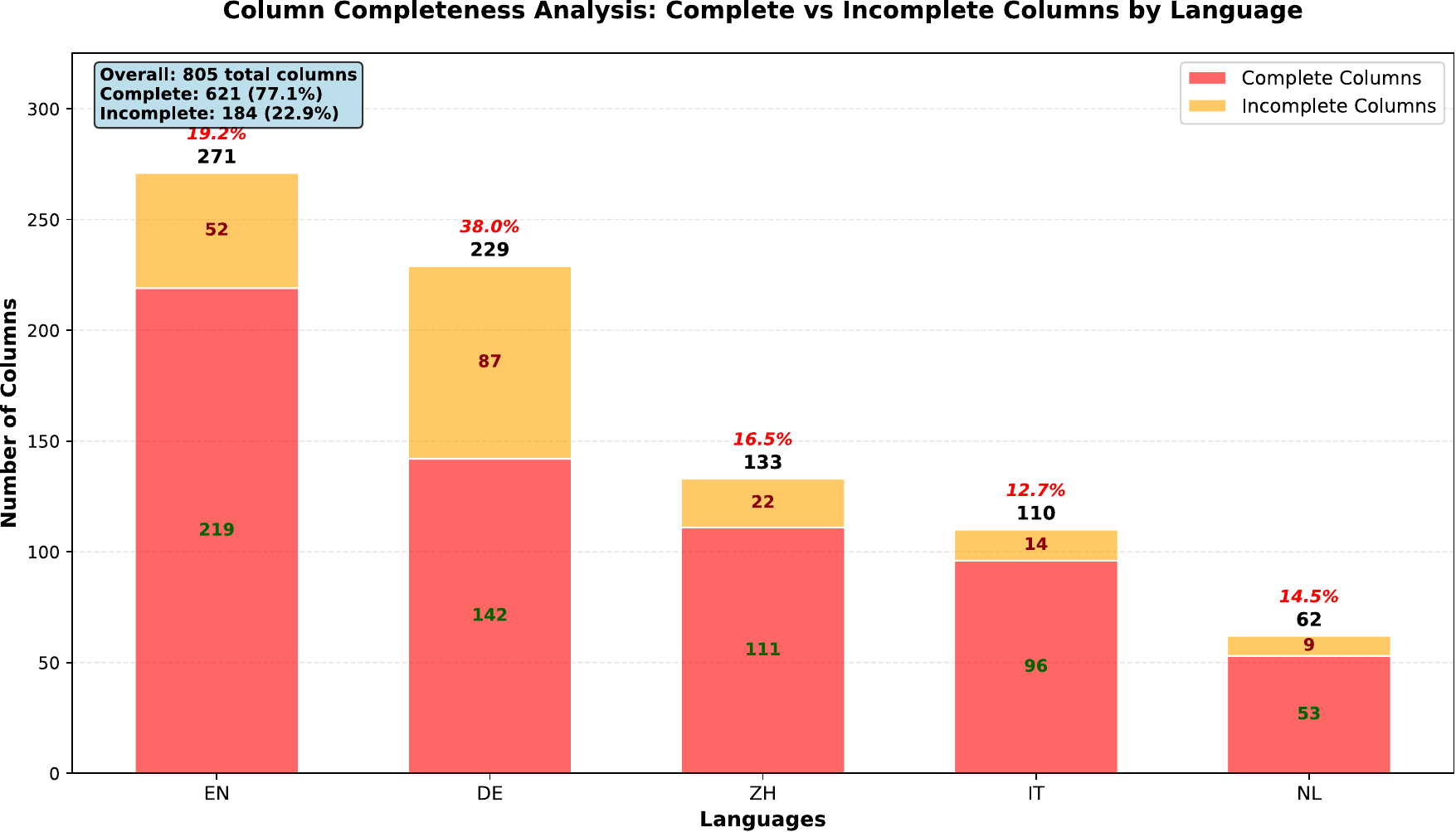}
    \caption{Information Completeness Analysis with Column Counts}
    \label{fig:column_count}
\end{figure}

\noindent
\textbf{Table Count:}
    The stacked bar chart in Figure~\ref{fig:table_numbers} reveals the distribution of Wikipedia tables across nine geographic and geological articles in five languages, with English (EN) dominating at 55 tables total, followed by German (DE) at 33 tables, Chinese (ZH) at 25 tables, Italian (IT) at 17 tables, and Dutch (NL) at 10 tables. While English maintains the largest contribution across most articles, German shows notable prominence in specific geographic topics, particularly excelling in mountain-related content such as "List of mountains of the Alps over 4000 metres" and "List of highest mountains on Earth," where it comprises a substantial portion of the available tables. Chinese Wikipedia, despite having a considerable number of tables where data exists, shows gaps in coverage with only six of the nine articles represented, indicating missing content for three entire articles. Similarly, both Italian and Dutch Wikipedia exhibit incomplete coverage across the articles and maintain fewer tables even in the articles they do cover, with Dutch having the most limited representation overall. This distribution pattern suggests that English Wikipedia serves as the most comprehensive resource for geographic and geological tabular data, while German Wikipedia demonstrates specialized prominence in mountain documentation, and the other languages show varying degrees of coverage gaps and content limitations.

\noindent
\textbf{Reference Count:}
    The heatmap in Figure~\ref{fig:reference_numbers} displays the distribution of references, with color intensity indicating reference density. 
    English Wikipedia demonstrates the highest reference count with an average of 94.6 references per article, significantly outperforming other language versions. Notable standouts in English include "Eight-thousander" with 264 references and "Lists of earthquakes" with 236 references, representing the most comprehensive documentation among all articles. German Wikipedia shows moderate reference activity with an average of 31.8 references per article, displaying relatively consistent coverage across most articles. Chinese Wikipedia presents a more selective pattern with an average of 17.4 references per article, showing concentrated effort in specific topics, most notably "Lakes of Titan" with 70 references, while having limited or no coverage for other articles. Italian Wikipedia maintains lower documentation with an average of 16.9 references per article, showing relatively balanced coverage across available articles. Dutch Wikipedia exhibits the most limited reference activity with an average of 4.9 references per article, with sparse coverage across most articles.

\noindent
\textbf{Column Count:}
    The bar chart in Figure~\ref{fig:column_count} displays column completeness analysis across the main tables from nine Wikipedia articles in five languages. This analysis focuses on the primary table from each Wikipedia page, as these represent the core information and are more significant than other tables. English Wikipedia demonstrates the highest volume with 271 total columns, of which 219 are complete and 52 are incomplete, resulting in a 19.2\% incompleteness rate. German Wikipedia shows 229 total columns but has the highest incompleteness rate at 38.0\%, with 87 incomplete columns out of 229 total, significantly reducing its complete column count to 142. In contrast, the other three languages exhibit much better data completeness: Chinese Wikipedia has 133 total columns with only 16.5\% incomplete, Italian Wikipedia shows 110 total columns with 12.7\% incomplete, and Dutch Wikipedia has 62 total columns with 14.5\% incomplete. Overall, across all 805 columns from the five language versions, 621 columns (77.1\%) are complete while 184 columns (22.9\%) contain missing or incomplete information, indicating that while English provides the most comprehensive coverage in terms of volume, German Wikipedia faces significant data quality challenges despite having substantial content.

\subsection{Qualitative Analysis}
\label{sec:qualitative-analysis}

To categorize various types of inconsistencies, we draw inspiration from the classification framework of knowledge deltas \cite{jarnac2024uncertainty}, and contextualize these categories with illustrative examples drawn from Wikipedia pages, we identify three potential sources of inconsistency that may affect it:

\noindent \textbf{Invalidity}: The value is incorrect or lacks credibility. As shown in Figure~\ref{fig:teaser-figure}, Wikipedia tables across different languages report conflicting death rate statistics for Mount Everest and K2. For example, the death rate for K2 is listed as 29.5\% in the Chinese version, 26.5\% in the Italian version, and inferred as 26.4\% in the German version (80 deaths out of 302 ascents), despite referencing similar or identical data. 
Such discrepancies highlight reliability issues in multilingual content consistency.

\begin{figure}[h!]
    \centering
    \vspace{-1em}
    \includegraphics[width=0.68\linewidth]{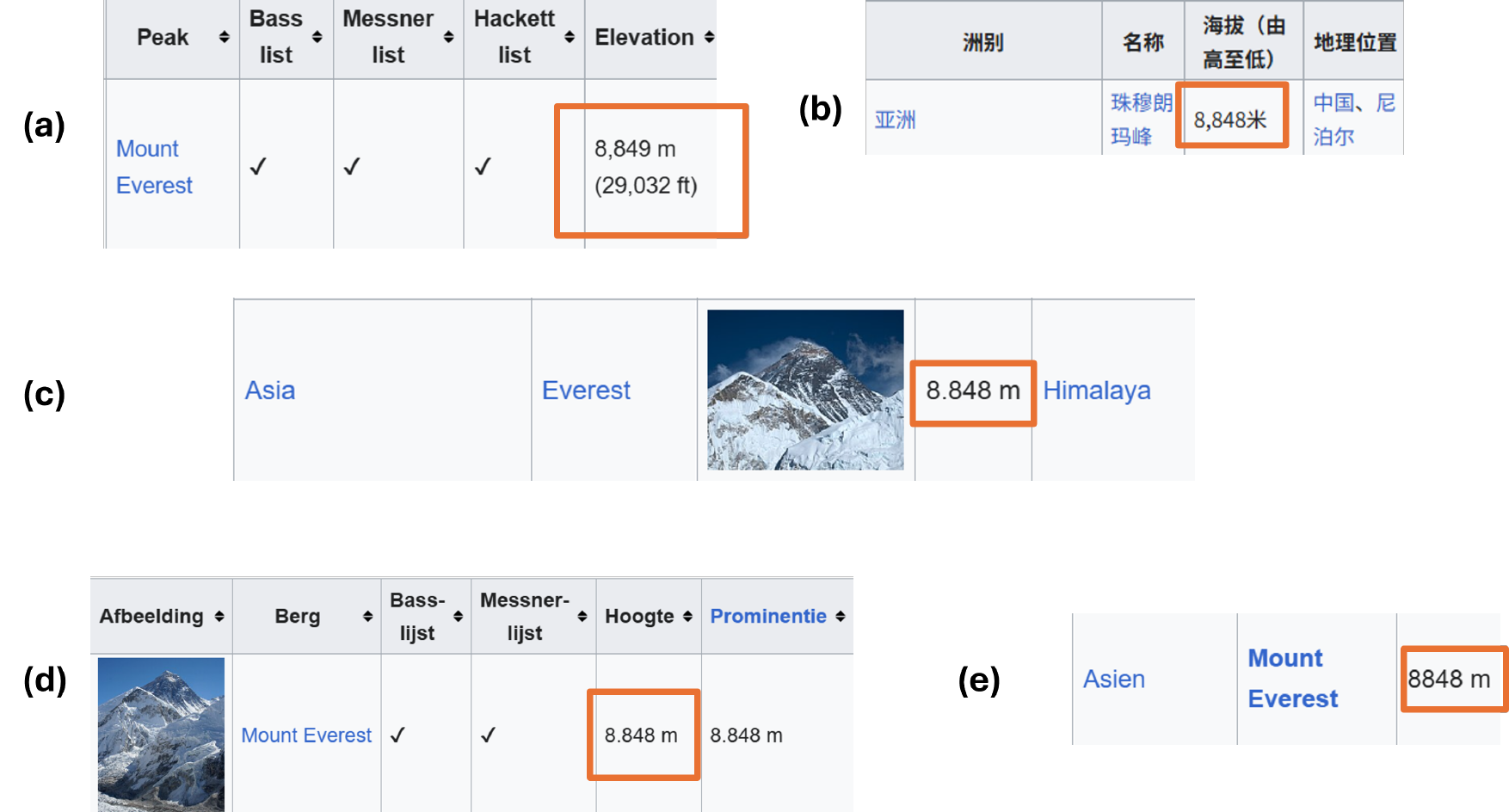}
    \vspace{-1em}
    \caption{Example of timeliness: height of Mount Everest differs across language versions of Wikipedia. The death rate of climbing the mountain is explicitly provided in (a) Italian and (b) Chinese. Only the absolute number of deaths is mentioned in (c) German. This information is absent in the English, Dutch, and Estonian versions.}
    \label{fig:example1}
\end{figure}

\noindent
\textbf{Timeliness}: A data source may present a statement \( t \) that was once valid but is now outdated, whereas another source may provide an updated version of \( t \). For example, in Figure~\ref{fig:example1}, the English version of the Wikipedia page states that Mount Everest's height is 8,849 meters, while another language version reports it as 8,848 meters. This discrepancy reflects a recent update, as  MountEverest has continued to grow—its height increasing by approximately 2mm per year—due to tectonic uplift and enhanced isostatic rebound triggered by erosion from river capture near the Arun River~\cite{Han2024Everest}.

\begin{figure}[h!]
    \centering
    \vspace{-1em}
    \includegraphics[width=0.68\linewidth]{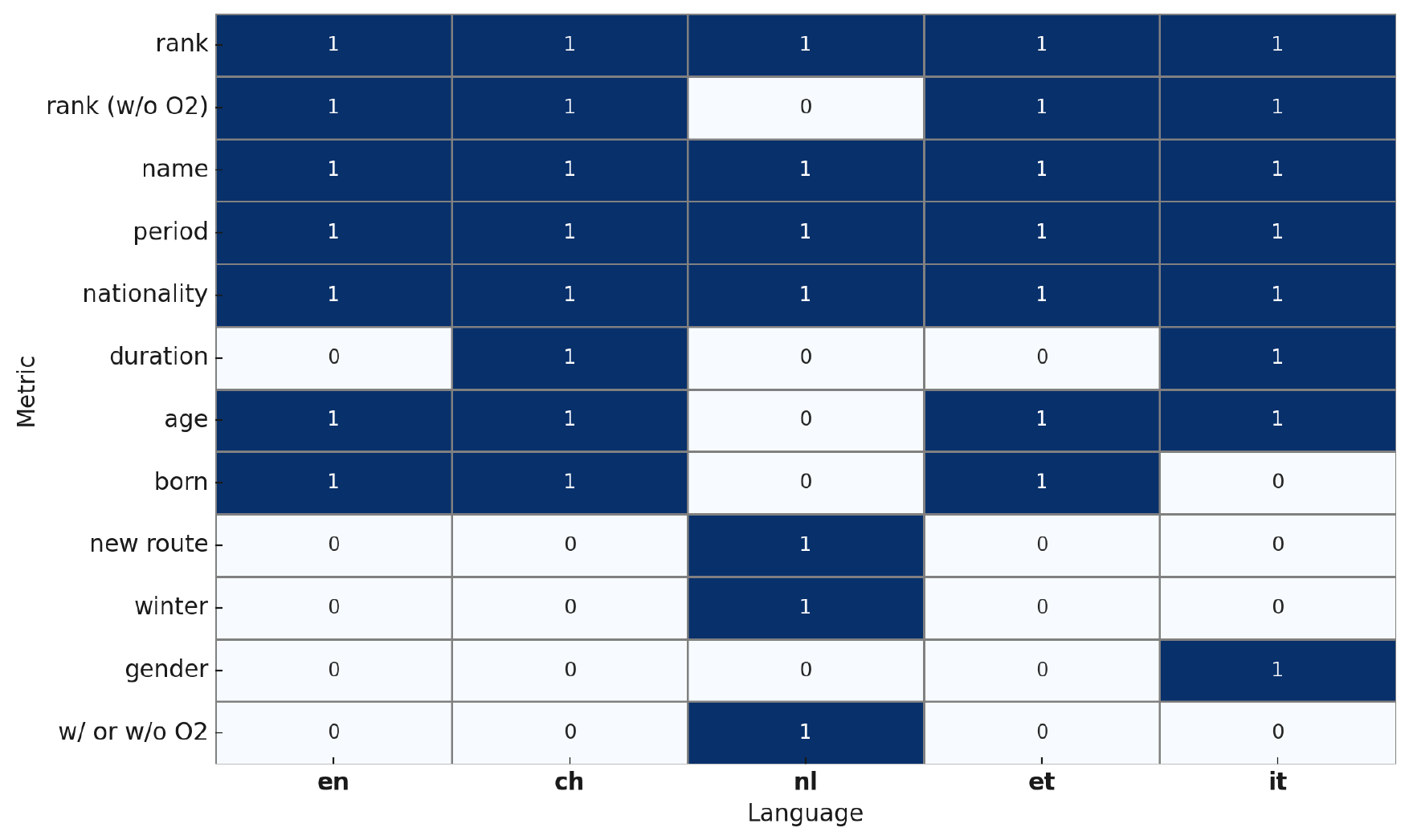}
    \vspace{-1em}
    \caption{Example of incompleteness: Binary heatmap of metric presence across five languages for the List of climbers who have summited all 14 eight-thousanders.}
    \label{fig:example3}
    
\end{figure}

\noindent
\textbf{Incompleteness}: One type of cross-lingual inconsistency we observe is schema-level incompleteness, where languages provide different sets of metrics (i.e., column headers) to describe the same underlying subject. 
Figure~\ref{fig:example3} illustrates this phenomenon using a binary heatmap that compares the metrics used in the Wikipedia tables for the List of climbers who have summited all 14 eight-thousanders across five language versions. While there is a clear overlap in core metrics such as rank, name, period, and nationality, several metrics are language-specific. For instance, the Dutch version includes unique metrics like new route and winter ascent, while the Italian version includes gender. On the other hand, some languages omit attributes found in others, where duration is present only in Chinese and Italian. This highlights the partial and uneven coverage of information across languages, even when describing the same real-world entities.

\section{Conclusions }
% Discussion about Knowledge Graphs for Reliable AI
% Future Works RQ4: How could such inconsistencies impact the reliability and fairness of AI systems (particularly LLMs) that rely on these knowledge sources?

In this work, we find that factual inconsistencies across language editions of Wikipedia include not only divergent values but also outdated data and missing or different structured content. We proposed a classification of these inconsistencies into three categories: Invalidity, Timeliness, and Incompleteness. We contextualized these categories with real-world examples from multilingual Wikipedia tables, highlighting how inconsistencies in statistical values, outdated information, and uneven schema coverage can all undermine the reliability of a knowledge. Since LLMs often rely on Wikipedia data, language discrepancies stemming from cultural differences can lead to biased knowledge representations and distortions in AI information processing, potentially disadvantaging certain users or perspectives. As for the underlying causes, we find that inconsistencies often arise from asynchronous updates across editions and from differences in source materials or cultural emphasis.
The integration of uncertain and multilingual knowledge continues to pose significant challenges. While progress has been made in representing uncertainty at both the ontological and data model levels, current approaches do not yet address the full diversity of inconsistencies.
%\textbf{Future Work}

For future work, we will focus on scaling the proposed analysis across a larger set of multilingual Wikipedia tables. This includes systematic selection of topic-aligned tables, aggregation of column headers, and cross-lingual embedding using models such as mT5, and XLM-RoBERTa. Pairwise cosine similarities will be computed to assess alignment, with annotated heatmaps used for visual analysis. Human evaluation will support semantic comparison of column headers to identify structural and editorial divergences. The results will refine inconsistency categories and reveal cultural influences on schema design across languages.

\vspace{.5cm}
\noindent
\textbf{GenAI Usage Disclosure.}
The authors used ChatGPT and Grammarly for limited assistance with writing polish and code debugging. All scientific content, including text, code, tables, figures, and claims, was authored by the authors.

\bibliographystyle{splncs04}
\bibliography{main}

\section{Appendix }

% ==== Figures ====

\end{document}